\documentclass[a4paper]{article}

\usepackage{INTERSPEECH2022}

\usepackage{multirow}
\usepackage{array,booktabs}
\usepackage{siunitx}
\usepackage{cite}
\usepackage{setspace}

\newcolumntype{P}[1]{>{\centering\arraybackslash}p{#1}}

 \newcommand{\textapprox}{\raisebox{0.5ex}{\texttildelow}}

\title{Effect and Analysis of Large-scale Language Model Rescoring \\on Competitive ASR Systems}
\name{Takuma Udagawa$^1$, Masayuki Suzuki$^1$, Gakuto Kurata$^1$, Nobuyasu Itoh$^1$, George Saon$^2$}
\address{
  $^1$IBM Research - Tokyo, Japan\\
  $^2$IBM T. J. Watson Research Center, Yorktown Heights, USA}
\email{Takuma.Udagawa@ibm.com, \{szuk,gakuto,iton\}@jp.ibm.com, gsaon@us.ibm.com}

\begin{document}

\maketitle
\begin{abstract}
Large-scale language models (LLMs) such as GPT-2, BERT and RoBERTa have been successfully applied to ASR N-best rescoring. However, whether or how they can benefit competitive, near state-of-the-art ASR systems remains unexplored. In this study, we incorporate LLM rescoring into one of the most competitive ASR baselines: the Conformer-Transducer model. We demonstrate that consistent improvement is achieved by the LLM's bidirectionality, pretraining, in-domain finetuning and context augmentation. Furthermore, our lexical analysis sheds light on how each of these components may be contributing to the ASR performance.

\end{abstract}

\section{Introduction}
\label{sec:introduction}

Large-scale language models (LLMs) such as GPT-2, BERT and RoBERTa \cite{radford2019language,devlin-etal-2019-bert,liu2019roberta} have become a prominent component in modern NLP. Based on their self-attention mechanism \cite{vaswani2017attention}, Transformer-based LLMs are capable of modeling long-range dependencies and interactions within the input, which are essential for higher-level language understanding. Conventionally, LLMs are first \textit{pretrained} on massive unlabelled text data, which allows them to learn the general linguistic knowledge (e.g. morphosyntax and semantics) \cite{tenney-etal-2019-bert,Tenney2019What} and world knowledge \cite{petroni-etal-2019-language}. Subsequently, these models can be \textit{finetuned} with a smaller amount of labelled data to achieve superior performance on specific (narrower) target tasks and domains \cite{ruder2021lmfine-tuning}.

To leverage the power of LLMs into ASR, recent work has proposed to rescore the ASR model's N-best hypotheses based on LLMs \cite{shin2019effective,salazar-etal-2020-masked,Li2020AnES,Chiu2021InnovativeBR,Zheng2021AdaptingGG,futami2021rescoring,xu2022rescorebert} in place of the traditional \textit{n}-gram or RNN-based LMs \cite{Rosenfeld2000TwoDO,Mikolov2010RecurrentNN,arisoy2015bidirectional}. In the most simple and foundational approach \cite{shin2019effective,salazar-etal-2020-masked}, LLM scores are computed as the \textit{log-likelihood} of each hypothesis for unidirectional models (e.g. GPT-2) and \textit{pseudo-log-likelihood} for bidirectional models (e.g. BERT and RoBERTa). By incorporating such LLM scores, we can expect to improve the linguistic acceptability of the final ASR results, including utterance-level consistency and discourse-level consistency across utterances \cite{xiong-etal-2018-session,Irie2019TrainingLM}.

However, existing empirical studies are conducted on modest ASR baselines with considerable room left for improvement, and it remains unclear whether or how LLM rescoring works in more competitive ASR systems. Therefore, in this study, we examine the effect of LLM rescoring on a strong Conformer-Transducer model \cite{Gulati2020Conformer}, which is a competitive, near state-of-the-art ASR baseline among non-attention-based models \cite{Tske2020SingleHA}. Furthermore, we assess how LLM rescoring improves ASR through a simple lexical analysis, where we decompose the error reduction rates based on word frequency and error type.

Based on our experiments with the widely used Switchboard dataset, we show that bidirectional LLMs (BERT and RoBERTa) consistently improve ASR results, while the unidirectional LLM (GPT-2) fails to do so on our competitive baseline. In addition, LLM's pretraining, in-domain finetuning and context augmentation (with past and future hypotheses) also consistently improves ASR. Finally, our lexical analysis sheds light on how each of these variants may be contributing to the ASR results, often providing complementary benefits.

\section{Methods}
\label{sec:methods}

In this work, we follow the most simple and foundational approach of LLM rescoring proposed in \cite{shin2019effective,salazar-etal-2020-masked}. The main idea is to compute the LLM score as the \textit{log-likelihood} for unidirectional models and \textit{pseudo-log-likelihood} for bidirectional models, as we will explain in Section \ref{subsec:unidirectional_llm_scoring} and \ref{subsec:bidirectional_llm_scoring}.

\subsection{Unidirectional LLM Scoring}
\label{subsec:unidirectional_llm_scoring}

Unidirectional language models, also known as \textit{causal language models}, are trained to predict the conditional probability of the next word given the prior history of words. To be precise, a unidirectional LLM with parameters $\Theta$ is trained to predict the probability $P_{\mathrm{LM}} (\bm{w}_t | \bm{W}_{< t}; \Theta)$, where $\bm{W}_{< t} := (\bm{w}_1,..., \bm{w}_{t-1})$ denotes the prior history of words up to $\bm{w}_t$. Then, a natural choice of the LLM score for the hypothesis $\bm{W} := (\bm{w}_1, \bm{w}_2,..., \bm{w}_{|\bm{W}|})$ would be its \textit{log-likelihood}, which can be computed by the chain rule of probability:
\begin{equation}
\label{eqn:causal_lm_score}
\mathrm{Score}_{\mathrm{LM}}(\bm{W}) := \sum^{|\bm{W}|}_{t=1} \log P_{\mathrm{LM}} (\bm{w}_t | \bm{W}_{< t}; \Theta)
\end{equation}

Note that such LLMs are \textit{unidirectional} in the sense that prediction of $\bm{w}_t$ only depends on the previous (left) context. One advantage of unidirectional LLM scoring is its computational efficiency, only requiring a single inference pass to compute the log-likelihood in an autoregressive manner.

\subsection{Bidirectional LLM Scoring}
\label{subsec:bidirectional_llm_scoring}

In contrast, bidirectional language models aim to predict $\bm{w}_t$ conditioned on both the left and right context. One representative example of bidirectional models are the \textit{masked language models}, which can be trained to estimate the conditional probability $P_{\mathrm{LM}} (\bm{w}_t | \bm{W}_{\backslash t}; \Theta)$, where $\bm{W}_{\backslash t} := (\bm{w}_1,..., \bm{w}_{t-1}, \texttt{[MASK]}, \bm{w}_{t+1}, ..., \bm{w}_{|\bm{W}|})$ denote the hypothesis $\bm{W}$ with the word $\bm{w}_t$ replaced by a special \texttt{[MASK]} token. Unfortunately, based on this likelihood, it is non-trival to estimate the exact log-likelihood of $\bm{W}$. Instead, prior work proposed to use its \textit{pseudo-log-likelihood} \cite{wang-cho-2019-bert} as the LLM score, which is given by the following equation:
\begin{equation}
\label{eqn:masked_lm_score}
\mathrm{Score}_{\mathrm{LM}}(\bm{W}) := \sum^{|\bm{W}|}_{t=1} \log P_{\mathrm{LM}} (\bm{w}_t | \bm{W}_{\backslash t}; \Theta)
\end{equation}

Note that by leveraging the \textit{bidirectional} (left and right) context, prediction of each $\bm{w}_t$ becomes more accurate, which can lead to more reliable LLM scoring. However, this comes with the cost of running the model $|\bm{W}|$ times (with each $\bm{w}_t$ replaced by the \texttt{[MASK]} token), which can be slow or computationally expensive. \\

Based on the LLM scores from equation (\ref{eqn:causal_lm_score}) or (\ref{eqn:masked_lm_score}), each hypothesis in the N-best can be rescored by the final score, which is a linear interpolation with the ASR model (AM) score:
\begin{equation}
\label{eqn:combined_score}
\mathrm{Score}(\bm{W}) := (1 - \lambda) \cdot \mathrm{Score}_{\mathrm{AM}}(\bm{W}) + \lambda \cdot \mathrm{Score}_{\mathrm{LM}}(\bm{W})
\end{equation}

\section{Experiments}
\label{sec:experiments}

\subsection{LLM Rescoring}
\label{subsec:llm_based_rescoring}

For LLM rescoring, we use the open-source GPT-2, BERT and RoBERTa from the huggingface library \cite{wolf-etal-2020-transformers}. While the pretrained models can be used out-of-the-box based on our method (Section \ref{sec:methods}), we also experiment with the following configurations to study the effect of LLM \textit{pretraining}, \textit{in-domain finetuning} and \textit{context augmentation}, respectively.

\textbf{Training from Scratch} (\textit{+scratch}):
To ablate the general knowledge acquired in the pretraining step, we train each LLM from scratch (i.e. using the same model architecture with randomly initialized parameters) based on the transcribed text of 2000 hours of Switchboard corpus.
%use the same model architecture and train each LLM from scratch

\textbf{In-domain Finetuning} (\textit{+finetune}):
While the pretraining corpora of LLMs cover diverse domains and topics, they mostly consist of written (non-spoken) language. To examine the additional gain from speech domain adaptation, we use the pretrained model parameters and finetune each LLM based on the same 2000 hours text of Switchboard corpus.

\textbf{Context Augmentation} (\textit{+context}):
Without changing the model parameters, we concatenate the 1-best of past and future hypotheses as additional context when computing the LLM scores \cite{Zheng2021AdaptingGG}. For the past hypotheses, the 1-best is obtained (and cached) after LLM rescoring with a fixed $\lambda$ value of 0.20 (from equation (\ref{eqn:combined_score})). For the future hypotheses, the 1-best is obtained from the initial ASR model. We use up to 40 tokens from the past 1-best and 20 tokens from the future 1-best, concatenated on the left and right of the current hypothesis.

Based on the results from our preliminary experiments, we insert special tokens at the beginning and ending (e.g. \texttt{[CLS]} and \texttt{[SEP]} for BERT) and do not indicate utterance boundaries (e.g. with a period \cite{Zheng2021AdaptingGG}) during context augmentation.  Focusing on the ceiling performance, we choose the best (and largest) $\lambda$ from $\{0.0, 0.05, ..., 1.00\}$ for equation (\ref{eqn:combined_score}) which achieve the lowest average WER on our test sets.

\subsection{ASR Baseline}
\label{subsec:asr_baseline}

To confirm the advantage of LLM rescoring, we trained a strong baseline neural transducer model.
%We used 262 hours of segmented speech from the standard 300-hour Switchboard-1 English conversational telephone speech as training data~\cite{george17:_englis_conver_telep_speec_recog_human_machin}. 
%We created 1300 hours of augmented training data by applying speed and tempo perturbation (90\% and 110\% for both perturbation) to the original 262 hours~\cite{ko2015audio}.
We used the widely used 1975 hours of English conversational telephone speech consisting of Switchboard, Fisher, and CallHome as training data\cite{george17:_englis_conver_telep_speec_recog_human_machin}.
We created 9875 hours of augmented training data by applying speed and tempo perturbation (90\% and 110\% for both perturbation) to the original 1975 hours~\cite{ko2015audio}.

For acoustic features, we used 240-dimensional features consisting of 40-dimensional log-Mel filterbank energies, their delta, and double-delta coefficients with frame stacking and a skipping rate of 2~\cite{sak2015fast}.
Training of the baseline model followed a recipe~\cite{saon21:_advan_rnn_trans_techn_speec_recog} that leverages various regularization techniques such as SpecAugment~\cite{park2019specaugment}, sequence noise injection~\cite{saon2019sequence}, DropConnect~\cite{wan2013regularization}, and speed and tempo perturbation~\cite{ko2015audio}. 
The Conformer-T model uses an encoder network of 10 Conformer blocks (512-dimensional feed forward module, 31-kernel convolution block, and 8 64-dimensional attention heads)~\cite{Gulati2020Conformer}. 
The encoder output is projected from a 512-dimensional acoustic embedding to 256 dimensions.
A prediction network uses a single unidirectional LSTM layer with 1024 cells followed by a projection from a 1024-dimensional linguistic embedding to 256 dimensions.
A joint network uses multiplicative integration and hyperbolic tangent, followed by a projection and softmax layer to represent 42 characters and an extra blank symbol. 
We used an efficient beam search algorithm, alignment-length synchronous decoding~\cite{saon20:_align_lengt_synch_decod_for_rnn_trans} for letter-based decoding, and used Word Error Rate (WER) as the evaluation metric.

Based on this Conformer-T model, we obtain the \textit{100-best hypotheses} searched with a beam size of 100 and a more realistic \textit{16-best hypotheses} searched with beam size 16. We report our baseline and LLM rescored results on the Hub5 2000 Switchboard (SWB) and CallHome (CH) evaluation test sets.

\section{Results}
\label{sec:results}

We show the LLM rescoring results with 16-best hypotheses in Table \ref{tab:conformer_16_best_results} and 100-best hypotheses in Table \ref{tab:conformer_100_best_results}. Results for the GPT-2 models are omitted since they did not improve WER ($0.1$ or more) with any value of $\lambda$ or model configurations.

% \begin{table}[th]
% \caption{
% Results of LLM rescoring.
% }
% \label{tab:rescoring_results}
% \centering \scalebox{0.97}{
% \setlength\tabcolsep{6pt}
% \begin{tabular}{lcccc}
% \toprule[\heavyrulewidth]
% \multirow{2}[3]{*}{Model} & \multicolumn{2}{c}{SWB} & \multicolumn{2}{c}{CH} \\
% \cmidrule(lr){2-3} \cmidrule(lr){4-5}
%  & \textit{16} & \textit{100} & \textit{16} & \textit{100} \\
% \midrule
% Baseline & 0 & 0 & 0 & 0 \\
% \midrule
% BERT (base, uncased) & 0 & 0 & 0 & 0 \\
% \phantom{00}+scratch & 0 & 0 & 0 & 0 \\
% \phantom{00}+finetune & 0 & 0 & 0 & 0 \\
% \phantom{00}+context & 0 & 0 & 0 & 0 \\
% \phantom{00}+finetune +context & 0 & 0 & 0 & 0 \\
% RoBERTa (base) & 0 & 0 & 0 & 0 \\
% \phantom{00}+scratch & 0 & 0 & 0 & 0 \\
% \phantom{00}+finetune & 0 & 0 & 0 & 0 \\
% \phantom{00}+context & 0 & 0 & 0 & 0 \\
% \phantom{00}+finetune +context & 0 & 0 & 0 & 0 \\
% \midrule
% BERT (large, uncased) & 0 & 0 & 0 & 0 \\
% \phantom{00}+scratch & 0 & 0 & 0 & 0 \\
% \phantom{00}+finetune & 0 & 0 & 0 & 0 \\
% \phantom{00}+context & 0 & 0 & 0 & 0\\
% \phantom{00}+finetune +context & 0 & 0 & 0 & 0 \\
% RoBERTa (large) & 0 & 0 & 0 & 0 \\
% \phantom{00}+scratch & 0 & 0 & 0 & 0 \\
% \phantom{00}+finetune & 0 & 0 & 0 & 0 \\
% \phantom{00}+context & 0 & 0 & 0 & 0 \\
% \phantom{00}+finetune +context & 0 & 0 & 0 & 0 \\
% \midrule
% Oracle & & & & \\
% \bottomrule[\heavyrulewidth]
% \end{tabular}
% }
% \end{table}

\begin{table}[th]
\caption{
Results of LLM rescoring (16-best hypotheses).
}
\label{tab:conformer_16_best_results}
\centering \scalebox{1.00}{
\setlength{\aboverulesep}{1pt}
\setlength{\belowrulesep}{1pt}
\setlength\tabcolsep{8pt}
\begin{tabular}{lccc}
\toprule[\heavyrulewidth]
Model & Best $\lambda$ & SWB & CH \\
\midrule
Baseline (Conformer-T) & - & 5.3 & 8.9 \\
\midrule
BERT-base (uncased) & 0.05 & 5.3 & 8.8 \\
\phantom{00}+scratch & 0.15 & 5.2 & 8.7 \\
\phantom{00}+finetune & 0.15 & \textbf{5.0} & \textbf{8.5} \\
\phantom{00}+context & 0.10 & 5.2 & 8.6 \\
\phantom{00}+finetune +context & 0.20 & \textbf{5.0} & \textbf{8.5} \\
\midrule
RoBERTa-base & 0.10 & 5.2 & 8.7 \\
\phantom{00}+scratch & 0.15 & 5.3 & 8.6 \\
\phantom{00}+finetune & 0.20 & \textbf{5.1} & 8.6 \\
\phantom{00}+context & 0.15 & 5.2 & 8.6\\
\phantom{00}+finetune +context & 0.20 & \textbf{5.1} & \textbf{8.5} \\
\midrule
BERT-large (uncased) & 0.05 & 5.2 & 8.8 \\
\phantom{00}+scratch & 0.15 & 5.2 & 8.7 \\
\phantom{00}+finetune & 0.25 & \textbf{5.1} & 8.5 \\
\phantom{00}+context & 0.10 & 5.2 & 8.6 \\
\phantom{00}+finetune +context & 0.15 & \textbf{5.1} & \textbf{8.4} \\
\midrule
RoBERTa-large & 0.10 & 5.1 & 8.6 \\
\phantom{00}+scratch & 0.10 & 5.2 & 8.6 \\
\phantom{00}+finetune & 0.25 & 5.2 & \textbf{8.5} \\
\phantom{00}+context & 0.10 & 5.1 & \textbf{8.5}\\
\phantom{00}+finetune +context & 0.15 & \textbf{5.0} & \textbf{8.5} \\
\midrule
Oracle (16-best) & - & 2.6 & 4.5 \\
\bottomrule[\heavyrulewidth]
\end{tabular}
}
\end{table}

\begin{table}[th]
\caption{
Results of LLM rescoring (100-best hypotheses).
}
\label{tab:conformer_100_best_results}
\centering \scalebox{1.00}{
\setlength{\aboverulesep}{1pt}
\setlength{\belowrulesep}{1pt}
\setlength\tabcolsep{8pt}
\begin{tabular}{lccc}
\toprule[\heavyrulewidth]
Model & Best $\lambda$ & SWB & CH \\
\midrule
Baseline (Conformer-T) & - & 5.3 & 8.9 \\
\midrule
BERT-base (uncased) & 0.05 & 5.3 & 8.8 \\
\phantom{00}+scratch & 0.10 & 5.2 & 8.7 \\
\phantom{00}+finetune & 0.15 & 5.0 & 8.5 \\
\phantom{00}+context & 0.10 & 5.2 & 8.5 \\
\phantom{00}+finetune +context & 0.15 & \textbf{4.9} & \textbf{8.4} \\
\midrule
RoBERTa-base & 0.10 & 5.2 & 8.6 \\
\phantom{00}+scratch & 0.15 & 5.3 & 8.6 \\
\phantom{00}+finetune & 0.15 & \textbf{5.0} & 8.5 \\
\phantom{00}+context & 0.15 & 5.2 & 8.5 \\
\phantom{00}+finetune +context & 0.15 & \textbf{5.0} & \textbf{8.4} \\
\midrule
BERT-large (uncased) & 0.10 & 5.2 & 8.7 \\
\phantom{00}+scratch & 0.15 & 5.2 & 8.7 \\
\phantom{00}+finetune & 0.20 & \textbf{5.1} & 8.4 \\
\phantom{00}+context & 0.10 & \textbf{5.1} & 8.6 \\
\phantom{00}+finetune +context & 0.15 & \textbf{5.1} & \textbf{8.3} \\
\midrule
RoBERTa-large & 0.10 & 5.1 & 8.6 \\
\phantom{00}+scratch & 0.10 & 5.2 & 8.6 \\
\phantom{00}+finetune & 0.15 & \textbf{5.0} & 8.5 \\
\phantom{00}+context & 0.15 & 5.1 & \textbf{8.4} \\
\phantom{00}+finetune +context & 0.15 & \textbf{5.0} & \textbf{8.4} \\
\midrule
Oracle (100-best) & - & 1.7 & 2.8 \\
\bottomrule[\heavyrulewidth]
\end{tabular}
}
\end{table}

Firstly, we note that LLM rescoring with 16-best hypotheses achieves absolute WER reduction of $0.3$ for SWB and $0.5$ for CH in the best case. With a larger budget of 100-best hypotheses, additional WER reduction of $0.1$ is achieved for both SWB and CH. This is a promising result for LLM-based rescoring with a very competitive ASR baseline.

We can also verify that rescoring with bidirectional LLMs work out-of-the-box using the pretrained parameters. While models trained from scratch are also generally effective (\textit{+scratch}), we observed consistent improvement when the models are rather finetuned from the pretrained parameters (\textit{+finetune}). Therefore, both general-domain pretraining and in-domain finetuning can be essential for LLM rescoring.

When context augmentation is conducted (\textit{+context}), we observed consistent improvement in both pretrained and finetuned LLMs; in fact, the latter achieves the best performance in all cases (\textit{+finetune} \textit{+context}). This result indicates that context augmentation has a complementary benefit with both pretraining and in-domain finetuning.

To further investigate the effect of context augmentation, we've also conducted the experiments with different context sizes, i.e. using the past (left) context only, future (right) context only, shorter ($\times \frac{1}{2}$) context size and longer ($\times 2$) context size. The results are shown in Table \ref{tab:different_context_size} (base-size models only).

From this experiment, we could verify that both the left and right context contribute to WER reduction. While longer ($\times 2$) context size improved rescoring in the case of BERT-base, we actually observed competitive performance with shorter ($\times \frac{1}{2}$) context size across all models and sizes (base and large). Therefore, we expect that local context (nearby \textapprox 20 words) has the highest impact on LLM rescoring.

%Lambda around 0.2 is good \cite{shin2019effective}.

In summary, bidirectional (but not unidirectional) LLM rescoring can be effective even on our competitive Conformer-T baseline, and LLMs benefit from all steps of general pretraining, in-domain finetuning and context augmentation.

\begin{table}[th]
\caption{
Context augmentation results with different context sizes (base-size models only, rescored with 100-best hypotheses).
}
\label{tab:different_context_size}
\centering \scalebox{1.00}{
\setlength{\aboverulesep}{1pt}
\setlength{\belowrulesep}{1pt}
\setlength\tabcolsep{4pt}
\begin{tabular}{lcP{1.0cm}P{1.0cm}cc}
\toprule[\heavyrulewidth]
\multirow{2}[3]{*}{Model} & \multicolumn{2}{c}{Context Size} & \multirow{2}[3]{*}{SWB} & \multirow{2}[3]{*}{CH} \\
\cmidrule{2-3}
& \# Left & \# Right &  &  \\
\midrule
\multirow{5}{*}{\shortstack[l]{BERT-base\\(uncased)\\+finetune}} & 40 & 20 & \textbf{4.9} & 8.4 \\
  & 40 & 0 & 5.0 & 8.5 \\
  & 0 & 20 & 5.0 & 8.5 \\
  & 20 & 10 & 5.0 & 8.4 \\
  & 80 & 40 & \textbf{4.9} & \textbf{8.3} \\
\midrule
\multirow{5}{*}{\shortstack[l]{RoBERTa-base\\+finetune}} & 40 & 20 & \textbf{5.0} & \textbf{8.4} \\
  & 40 & 0 & 5.1 & \textbf{8.4} \\
  & 0 & 20 & \textbf{5.0} & 8.5 \\
  & 20 & 10 & \textbf{5.0} & \textbf{8.4} \\
  & 80 & 40 & \textbf{5.0} & \textbf{8.4} \\
% \midrule
%   \multirow{5}{*}{\shortstack[l]{BERT-large\\(uncased)\\+finetune}} & 40 & 20 & 5.1 & \textbf{8.3} \\
%   & 40 & 0 & 5.1 & 8.4 \\
%   & 0 & 20 & 5.1 & 8.4 \\
%   & 20 & 10 & 5.1 & \textbf{8.3} \\
%   & 80 & 40 & \textbf{5.0} & 8.4 \\
% \midrule
% \multirow{5}{*}{\shortstack[l]{RoBERTa-large\\+finetune}} & 40 & 20 & \textbf{5.0} & 8.4 \\
%   & 40 & 0 & \textbf{5.0} & 8.5 \\
%   & 0 & 20 & 5.1 & \textbf{8.3} \\
%   & 20 & 10 & \textbf{5.0} & 8.4 \\
%   & 80 & 40 & 5.1 & \textbf{8.3} \\
\bottomrule[\heavyrulewidth]
\end{tabular}
}
\end{table}

\section{Analysis and Discussions}
\label{sec:analysis_and_discussions}

\subsection{Lexical Analysis}
\label{subsec:lexical_analysis}

In this section, we conduct a further analysis on how LLM rescoring improves ASR on our competitive baseline. In prior work, it has been reported that bidirectional LLM rescoring improves ASR in short utterances or at the earlier position of the utterances \cite{shin2019effective}. It's also reported that bidirectional LLMs are better at judging the linguistic acceptability (e.g. grammaticality) across a wide range of English phenomena \cite{salazar-etal-2020-masked,warstadt-etal-2020-blimp-benchmark}.

In contrast to prior work, we conduct a simple \textit{lexical analysis} based on the word frequency and error type. To be specific, we first classify each ASR error into the following word frequency classes (\textit{high}, \textit{medium} and \textit{low}) defined based on the transcribed text of 2000 hours Switchboard corpus:
%To be specific, we compare the error reduction rates for each error type (\textit{deletion} and \textit{insertion}) and word frequency class (\textit{high}, \textit{medium} and \textit{low frequency}) which is defined based on the transcribed text of 2000 hours Switchboard corpus as follows:

% In contrast, we conduct a simple \textit{lexical analysis} based on the word frequency and error type. First, we define the following word frequency class (\textit{high}, \textit{medium} and \textit{low}) based on the transcribed text of 2000 hours Switchboard corpus:
%To be specific, we first classify each ASR error into the following word frequency class (\textit{high}, \textit{medium} and \textit{low frequency}) defined based on the transcribed text of 2000 hours Switchboard corpus:

\noindent
\textbf{High Frequency}: The set of words $w$ whose unigram probability is larger than 0.1 ($P_{\mathrm{UNI}}(w) > 0.1$) on Switchboard. This class includes common pronouns, fillers and function words: e.g. \textit{i}, \textit{you}, \textit{who}, \textit{um}, \textit{well}, \textit{guess}, \textit{could}, \textit{to}, \textit{not}, \textit{when}.

\noindent
\textbf{Medium Frequency}: The set of words $w$ whose unigram probability is $0.0001 < P_{\mathrm{UNI}}(w) \leq 0.1$ on Switchboard. This class mainly includes the content words: e.g. \textit{listen}, \textit{east}, \textit{women}, \textit{summer}, \textit{brother}, \textit{normal}, \textit{laugh}, \textit{initially}.

\noindent
\textbf{Low Frequency}: The set of low frequency words that do not belong to either high or medium frequency class. This class includes rare words and named entities: e.g.  \textit{firestone}, \textit{realtor}, \textit{threshold}, \textit{worldly}, \textit{unprofitable}, \textit{whatev}, \textit{lehigh}.

The high frequency class contains 138 words and comprises 70.5\% of the Switchboard corpus, while the medium frequency class contains 12,049 words and comprises 28.7\% of Switchboard. The low frequency class is long-tailed ($\geq$ 35,602 words) but only comprises 0.8\% of the corpus.

In addition, we classify each error based on its error type, namely \textit{deletion} and \textit{insertion}. Note that we count a \textit{substitution} error (e.g. \textit{``tried''} $\rightarrow$ \textit{``trade''}) as one deletion error (of \textit{``tried''}) and one insertion error (of \textit{``trade''}) in our analysis.

Based on this classification, we compute the relative error reduction rates against our Conformer-T baseline, separately for each word frequency class and error type.

\subsection{Analysis Results}
\label{subsec:analysis_results}

\begin{table*}[th]
\caption{
Relative error reduction rates against our Conformer-T baseline (rescored with 100-best hypotheses, \underline{higher $\uparrow$ is better}). The results are classified based on word frequency (\textbf{high}, \textbf{medium}, and \textbf{low}) and error type (\textbf{Del.} for deletion and \textbf{Ins.} for insertion). We also report the overall error reduction rate (\textbf{Overall}) including both error types and mark the best results in bold.
}
\label{tab:analysis_llm_rescoring}
\centering \scalebox{1.00}{
\setlength{\aboverulesep}{1pt}
\setlength{\belowrulesep}{1pt}
\setlength\tabcolsep{7pt}
\begin{tabular}{lccccccccc}
\toprule[\heavyrulewidth]
 & \multicolumn{3}{c}{High Freq.} & \multicolumn{3}{c}{Medium Freq.} & \multicolumn{3}{c}{Low Freq.} \\
\cmidrule(lr){2-4} \cmidrule(lr){5-7} \cmidrule(lr){8-10}
 &  Del. & Ins. & Overall & Del. & Ins. & Overall & Del. & Ins. & Overall \\
\midrule
BERT-base (uncased) & 0.6 & $-$0.5 & 0.1 & 2.9 & 4.1 & 3.5 & 2.0 & 7.9 & 4.3 \\
\phantom{00}+scratch & 1.3 & $-$0.4 & 0.5 & 4.9 & 5.8 & 5.4 & 3.0 & 18.7 & 9.1 \\
\phantom{00}+finetune & 4.5 & 1.8 & 3.2 & 10.3 & 13.4 & 11.8 & 4.8 & 22.6 & 11.7 \\
\phantom{00}+context & 1.6 & 3.1 & 2.3 & 7.1 & 10.6 & 8.8 & 5.5 & 15.1 & 9.2 \\
\phantom{00}+finetune +context & 5.8 & 3.1 & \textbf{4.4} & 10.7 & 14.1 & \textbf{12.4} & 5.5 & 23.4 & \textbf{12.4} \\
\midrule
RoBERTa-base & 2.2 & 1.4 & 1.8 & 7.0 & 9.4 & 8.2 & 4.3 & 17.9 & 9.5 \\
\phantom{00}+scratch & 1.1 & 0.4 & 0.8 & 5.6 & 3.1 & 4.4 & 2.3 & 31.0 & 13.4 \\
\phantom{00}+finetune & 3.5 & 2.5 & 3.0 & 10.0 & 11.7 & 10.8 & 6.0 & 25.4 & \textbf{13.5} \\
\phantom{00}+context & 0.3 & 4.7 & 2.5 & 7.9 & 12.3 & 10.1 & 7.8 & 21.4 & 13.1 \\
\phantom{00}+finetune +context & 4.7 & 3.7 & \textbf{4.2} & 10.7 & 11.4 & \textbf{11.0} & 4.8 & 25.4 & 12.7 \\
% \midrule
% BERT-large (uncased) & 0.9 & 0.1 & 0.5 & 6.7 & 8.1 & 7.4 & 7.0 & 17.1 & 10.9 \\
% \phantom{00}+scratch & 1.7 & $-$2.2 & $-$0.3 & 6.3 & 7.0 & 6.6 & 3.0 & 27.4 & 12.4 \\
% \phantom{00}+finetune & 3.9 & 1.9 & 2.9 & 9.6 & 14.6 & 12.0 & 7.0 & 29.0 & 15.5 \\
% \phantom{00}+context & 1.7 & 2.9 & 2.3 & 8.1 & 11.7 & 9.9 & 5.5 & 13.5 & 8.6 \\
% \phantom{00}+finetune +context & 5.9 & 1.5 & 3.7 & 11.4 & 13.0 & 12.2 & 4.3 & 27.0 & 13.1 \\
% \midrule
% RoBERTa-large & 2.3 & 3.5 & 2.9 & 6.8 & 6.4 & 6.6 & 5.8 & 19.0 & 10.9 \\
% \phantom{00}+scratch & 1.6 & 0.4 & 1.0 & 6.6 & 5.1 & 5.9 & 4.3 & 23.8 & 11.8 \\
% \phantom{00}+finetune & 5.1 & 2.9 & 4.0 & 7.9 & 9.3 & 8.6 & 5.8 & 27.0 & 14.0 \\
% \phantom{00}+context & 1.4 & 6.3 & 3.8 & 9.3 & 14.8 & 12.0 & 7.3 & 23.4 & 13.5 \\
% \phantom{00}+finetune +context & 5.0 & 3.0 & 4.0 & 11.1 & 11.7 & 11.4 & 3.8 & 25.0 & 12.0 \\
\midrule
GPT-2 & $-$5.9 & 7.0 & \textbf{0.4} & 1.8 & $-$1.1 & \textbf{0.3} & 0.0 & 7.1 & 2.8 \\
\phantom{00}+finetune & $-$4.5 & 5.0 & 0.1 & 2.6 & $-$2.0 & \textbf{0.3} & $-$2.3 & 12.3 & 3.4 \\
\phantom{00}+finetune +context & $-$4.5 & 4.7 & 0.0 & 2.5 & $-$2.0 & \textbf{0.3} & $-$1.3 & 13.5 & \textbf{4.5} \\
\bottomrule[\heavyrulewidth]
\end{tabular}
}
\end{table*}

We summarize the results of our lexical analysis in Table \ref{tab:analysis_llm_rescoring} (base-size models only). We also report the overall error reduction rates (including both deletion and insertion errors)\footnote{Note that the overall reduction rate is equivalent to the \textit{weighted average} of deletion and insertion error reduction rates.} and mark the best result for each word frequency class in bold.

%Note that \textit{both} is identical to the weighted average of deletion and insertion error reduction.

First of all, in general we notice a trade-off between deletion and insertion error reduction, i.e. when deletion error is reduced (correct words are inserted), insertion error tends to increase (incorrect words are inserted). Secondly, there is a significant difference in the results between bidirectional LLMs (BERT and RoBERTa) and unidirectional LLMs (GPT-2), which we discuss separately in the following.
%insertion error increases (incorrect words are inserted), deletion error tends to decrease (correct words are inserted).

Starting with bidirectional LLMs, we found that the error reduction rates are consistently higher for medium and low frequency word classes. In the low frequency class, the reduction rate for the insertion error is especially high and conspicuous. Therefore, while LLM also improves on high frequency words, LLM's knowledge is most helpful for improving ASR on content words and \textit{removing} incorrect rare words.

Next, if we focus on the results with finetuning (\textit{+finetune}), we can verify that the improvement over scratch training (\textit{+scratch}) is visible in almost all error types and word frequency classes. Therefore, pretraining seems to have a positive effect in general with no (or minimal) side effects.

We can also observe that context augmentation (\textit{+context}) has a benefit similar to finetuning, but there is one interesting difference in the high frequency class: finetuning helps more in reducing deletion errors (i.e. inserting correct words), while context augmentation helps more on the insertion errors (i.e. deleting incorrectly inserted words). This difference seems to have a complementary benefit, and the overall error reduction is most significant (in terms of relative improvement) with the high frequency class when finetuning and context augmentation are combined: e.g. 37.5\% improvement ($3.2$ $\rightarrow$ $4.4$) for BERT-base with \textit{+finetune} $\rightarrow$ \textit{+finetune} \textit{+context}.

Finally, in the unidirectional LLM (GPT-2), we found that the model tends to have an unbalanced effect, improving one type of error but making it worse on the other. For instance, in the high frequency class, insertion errors are significantly reduced in exchange for the increase in deletion errors; this indicates the model is \textit{over-deleting} high frequency words. Such failure modes were not remedied with either finetuning or context augmentation, and we expect that the root of the cause is the lack of bidirectional (left and right) context which is necessary for reliable and effective LLM scoring.

% We expect that high frequency words are difficult to predict only with the past (left) context, and the failure in the unidirectional LLM

% this is due to the insufficiency of the context, 

% the insufficiency of the context 

%  which may be difficult to predict only with the past (left) context. 
%This is most likely due to the insufficiency of the context.

\section{Related Work}
\label{sec:related_work}

As a related work, there have been several attempts to improve the original LLM rescoring method of \cite{shin2019effective,salazar-etal-2020-masked}. For instance, the N-best can be directly rescored based on discriminative training with LLM to optimize for WER reduction \cite{Chiu2021InnovativeBR,futami2021rescoring,xu2022rescorebert}. Other works proposed improvements on bidirectional LLM rescoring, e.g. reducing computational cost by predicting the pseudo-log-likelihood based on regression \cite{salazar-etal-2020-masked,xu2022rescorebert} or seeking for the exact log-likelihood through a recursive decomposition of pseudo-log-likelihood \cite{Zheng2021AdaptingGG}. In contrast, the goal of our study is not to propose such improvements but to examine the effect of the original LLM rescoring on a competitive baseline.

Several works also proposed the approach of \textit{knowledge distillation} instead of N-best rescoring to infuse the power of LLMs \cite{Futami2020DistillingTK,kubo2022knowledge}. One advantage of N-best rescoring is that it requires no modification to the ASR models, allowing for a fast experiment turnover and analysis.

% Recursive procedure to compute the exact language prior probability \cite{Zheng2021AdaptingGG}.

% Knowledge distillation \cite{Futami2020DistillingTK}

% Existing works use non-sota Baselines:

% Bidirectional long-short term memory (BLSTMP) model \cite{shin2019effective,salazar-etal-2020-masked}

% While there has been recent attempts to improve this rescoring method 

% With a well-configured causal LLMs such as GPT-2, previous work reports improved ASR results \cite{Irie2019LanguageMW}.

% (Need rephrasing)
% In N-best rescoring, ASR modules remain unchanged while having a fast experiment turnover.
% we can avoid any modification of the modules of an ASR system and thus have fast experimental turnover.

% Related work:
% optimize for word error rate (rather than token prediction probability), e.g. based on discriminative training
% more accurate scoring?
% faster inference for masked language models

% Recently, there has been several attempts to improve this rescoring method, e.g. to optimize for the WER reduction, reduce the computational cost of pseudo-log-likelihood, or estimate the exact log-likelihood of masked language models.

\section{Conclusions}

In this study, we have re-examined the effect of the fundamental LLM rescoring approach \cite{shin2019effective,salazar-etal-2020-masked} on a competitive Conformer-Transducer baseline and conducted a detailed analysis.
Based on our experiments, we have demonstrated consistent improvement in ASR accuracy using bidirectional (but not unidirectional) LLM rescoring. We also observed additional gains from general-domain pretraining, in-domain finetuning and context augmentation when using the bidirectional LLMs.

Lastly, we've conducted a simple lexical-based analysis to examine the effect of LLM rescoring. We showed that error reduction from rescoring can be different across (and characterized by) the word frequency and error type. 
%Based on our analysis, we shed light on how each component of LLM rescoring contributes to ASR accuracy improvement and explain the failure mode of unidirectional LLMs, being unable to balance both error types.
Based on our analysis, we shed light on how each variant of LLM contributes to WER reduction and explain the failure mode of unidirectional LLMs, being unable to balance both error types.
%Our analysis also sheds light on how each configuration of LLM rescoring improves ASR and explains the failure mode of unidirectional LLMs.

\newpage

\bibliographystyle{IEEEtran}

\bibliography{mybib}

% \begin{thebibliography}{9}
% \bibitem[1]{Davis80-COP}
%   S.\ B.\ Davis and P.\ Mermelstein,
%   ``Comparison of parametric representation for monosyllabic word recognition in continuously spoken sentences,''
%   \textit{IEEE Transactions on Acoustics, Speech and Signal Processing}, vol.~28, no.~4, pp.~357--366, 1980.
% \bibitem[2]{Rabiner89-ATO}
%   L.\ R.\ Rabiner,
%   ``A tutorial on hidden Markov models and selected applications in speech recognition,''
%   \textit{Proceedings of the IEEE}, vol.~77, no.~2, pp.~257-286, 1989.
% \bibitem[3]{Hastie09-TEO}
%   T.\ Hastie, R.\ Tibshirani, and J.\ Friedman,
%   \textit{The Elements of Statistical Learning -- Data Mining, Inference, and Prediction}.
%   New York: Springer, 2009.
% \bibitem[4]{YourName17-XXX}
%   F.\ Lastname1, F.\ Lastname2, and F.\ Lastname3,
%   ``Title of your INTERSPEECH 2022 publication,''
%   in \textit{Interspeech 2022 -- 23\textsuperscript{rd} Annual Conference of the International Speech Communication Association, September 18-22, Incheon, Korea, Proceedings, Proceedings}, 2022, pp.~100--104.
% \end{thebibliography}

\end{document}